# Pose-Invariant 3D Face Reconstruction

Lei Jiang[1,3], Xiao-Jun Wu[1,3,*], Josef Kittler[2]

[1] School of IoT Engineering, Jiangnan University, 214122, Wuxi, China

[2] Center for Vision, Speech and Signal Processing(CVSSP), University of Surry, GU2 7XH, Guildford, UK

[3] Jiangsu Provincial Engineering Laboratory of Pattern Recognition and Computational Intelligence, Jiangnan University, 214122, Wuxi, China.

ljiang_jnu@outlook.com   xiaojun_wu_jnu@163.com   j.kittler@surrey.ac.uk

## Abstract

*3D face reconstruction is an important task in the field of computer vision. Although 3D face reconstruction has being developing rapidly in recent years, it is still a challenge for face reconstruction under large pose. That is because much of the information about a face in a large pose will be unknowable. In order to address this issue, this paper proposes a novel 3D face reconstruction algorithm (PIFR) based on 3D Morphable Model (3DMM). After input a single face image, it generates a frontal image by normalizing the image. Then we set weighted sum of the 3D parameters of the two images. Our method solves the problem of face reconstruction of a single image of a traditional method in a large pose, works on arbitrary Pose and Expressions, greatly improves the accuracy of reconstruction. Experiments on the challenging AFW, LFPW and AFLW database show that our algorithm significantly improves the accuracy of 3D face reconstruction even under extreme poses (±90 yaw angles).*

## 1. Introduction

This paper aims to advance face reconstruction from a single face images with arbitrary poses. 3D face reconstruction is the problem of recovering the 3D facial geometry from 2D images image. It arises in many applications, such as 3D emoticon animation [1,2], face recognition [3], 3D face Alignment [4,5,6], face landmark detection [7,8], etc. Although it has developed very rapidly in recent years, it is still an open topic and has a high degree of attention. The 3D morphable face model(3DMM) proposed by Blanz and Vetter in 1999 [9] uses a laser scanner to obtain a 3D prototype face, and uses a prototype face to create a face deformation model, and the model is matched to a 2D face image to realize 3D reconstruction of the face. The 3D morphable model considers the pose and illumination of the face when expressing the face image, so the representation of the face image is better. In addition, since the model considers the 3D modeling problem of the face as the error optimization of the model and the given face image, it can be solved by using the optimization method to realize the specific automatic face modeling based on a single image.

3D morphable model adopts the idea of linear combination to construct the basic face space, and adjusts the coefficient to determine the expression of the specific 3D dimensional face model. Therefore, a 3D dimensional face model can be obtained by calculating relevant 3D parameters. In recent years, some innovative methods have been proposed, such as estimating 3D parameters through CNN networks [5,10,11,12] and related cascaded regression [6,12] operations to achieve 3D face reconstruction. At the same time, more methods use face feature point operations instead of dense correspondence. In the traditional method, the least square method is used to estimate the 3D parameters. Since the landmark of different parts of the face has different semantic information, [12] uses the weighted landmark to calculate the 3D model of the face. However, the performance of face reconstruction in large pose is significantly degraded, mainly because many of the face information will be lost in the case of large pose, and some facial feature points will be invisible.

In this paper, we propose PIFR, which mainly solves the problem of reconstruction of traditional methods in large poses. An overview of our method is shown in Figure.1. We discuss the proposed method in more details in Sec. 3. In summary, our contributions are summarized as follows:

1) By normalizing the input image, more identity information of the face and visibility of the face feature points are restored.
2) The weighted fusion of the front image and the original image not only preserves the pose of the original image, but also enhances the identity information.
3) We experimentally verified that our algorithm has significantly improved performance of 3D face reconstruction compared to the previous algorithms, and supports the face reconstruction problem with extreme pose.

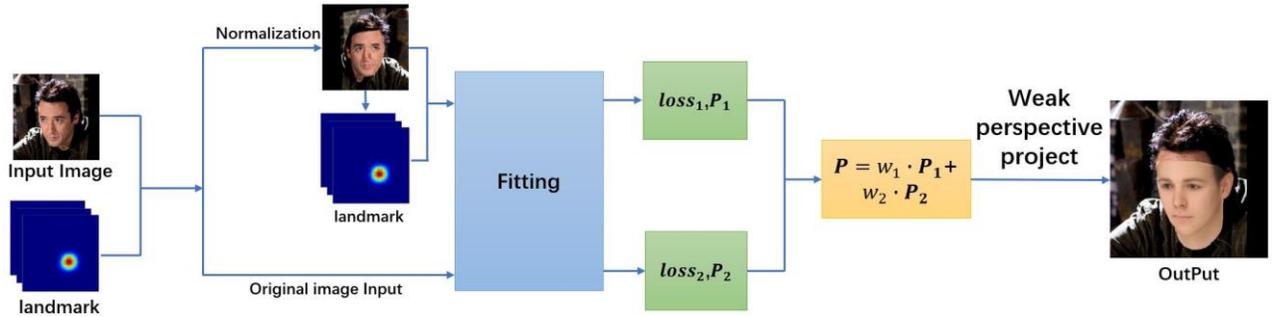

Figure 1. Overview of the Pose-Invariant 3D Face Reconstruction (PIFR) method

The paper is organised as follows. Section 2 presents a brief review of the related literature. Section 3. Our method is described in detail. The advocated approach is validated experimentally in Section 4 and the paper is drawn to conclusion in Section 5.

## 2. Related Work

In this section, we will review the work on 3D face models and 3D face reconstruction.

### 2.1. 3D Face Models

The 3D Morphable Model is a typical statistical 3D face model. It has a clear understanding of the prior knowledge of 3D faces through statistical analysis. It means that a specific three-dimensional face is a linear combination of basic three-dimensional faces, obtained by principal component analysis (PCA) on a group of densely arranged 3D faces. Considering the 3D face reconstruction problem as a model fitting problem, the model parameters (ie linear combination coefficients and camera parameters) are optimized to produce a 3D face that is projected through the perspective to best fit the position (and texture) of the input 2D image, And a set of facial feature point markers (eg, eye center, mouth corner, and nose tip). 3DMM-based methods usually require online optimization, so they are computationally intensive, so the real-time performance is poor. In addition, it should be noted that PCA is essentially a low-pass filter, so this method is unsatisfactory in restoring the detailed features of the face. Although this model has the above problems, it is still widely used in the field of computer vision such as face recognition [3,13], face alignment [14], face animation [15], etc.

There are currently four publicly available three-dimensional deformation models. First, the BFM [16] model proposed by the University of Basel. Second, the 3DMM models developed by Bolkart et.al [17], Which model 3D face shapes of different subjects in neutral. Third, a multi-resolution 3D face model provided by University of Surrey, UK [18]. Fourth, the large-scale face model built by Imperial College LSFM [19], which contains faces of different races and ages. The most widely used BFM model is used in this paper.

### 2.2. 3D Face Reconstruction

In order to improve the face recognition accuracy under different working conditions such as different illumination and different angles, the 3D face model can be reconstructed from 2D face, and more face data of different angles can be obtained for training, thereby improving face recognition accuracy. In addition, using 3D face data for face recognition is more robust and more accurate than using 2D face images. Especially in the complex situation of large face angle, ambient light change, make-up, and expression changes, it still has high recognition accuracy. Because the 3D face contains spatial information about the face relative to the 2D face image data. However, high-resolution and high-precision 3D face data is not so easy to obtain, especially under practical working conditions such as various complicated or long-distance shooting. 2D face data is relatively easy to obtain, so how to better reconstruct 3D face model through 2D face image is an important direction to explore in face recognition.

The work of [12] uses a cascaded regression method to reconstruct a 3D model based on 2D face images. Because the basic regression method also has the advantages of the regression method, the effect is better and the speed is faster. At the same time, for the case where the key points of the face are missing due to the pose, the displacement deviation of the key points is uniformly represented by a constant such as zero, so that the missing part of the key point can be processed. Tulyakov S et al. [20] puts the face key point detection and 3D face reconstruction together. The author proposed 3d-shape-feature-indexing to construct the tree-based regression model, and also introduced the 3D model based pose estimation to improve the final Reconstruction accuracy. Ref [21] proposes a 3D morphable model based Pose and Expression normalization method, which can automatically generate frontal and normal expression faces according to faces of different Pose and Expressions, thereby improving face recognition accuracy. In order to express the variation of face expression, the extended 3DMM(E-3DMM) [22] was suggested. Based on the traditional regression method, the paper [30] performs a weighted landmark 3DMM fitting, which has achieved great improvement in small and medium pose. Our method has varying degrees of performance improvement in arbitrarily pose. In particular, the performance improvement is more significant in large posture.

## 3. Pose-Invariant 3D Face Reconstruction

In this section we mainly introduce some details of our proposed Pose-Invariant 3D Face Reconstruction (PIFR). We will describe each part of the method and the reasons in detail.

### 3.1. 3D Morphable Model

The 3D Morphable model is one of the most successful methods for describing 3D face space. Blanz et al. [23] proposed a 3D morphable model (3DMM) of 3D face space with PCA. It is expressed as follows:

$$S = \bar{S} + A_{id}\alpha_{id} + A_{exp}\alpha_{exp} \quad (1)$$

where $S$ is a specific 3D face, $\bar{S}$ is the mean face, $A_{id}$ is the principle axes trained on the 3D face scans with neutral expression and $\alpha_{id}$ is the shape parameter, $A_{exp}$ between expression scans and neutral scans and $\alpha_{exp}$ is the expression parameter. So the coefficient $\{\alpha_{id}, \alpha_{exp}\}$ defines a unique 3D face. In this work, $A_{id}$ come from the BFM model and $A_{exp}$ come from the FaceWarehouse model [24]. a unique 3D face and its 3D landmark is shown in Figure.2.

In the process of 3DMM fitting, we use the Weak Perspective Projection to project 3DMM onto the 2D face plane. This process can be expressed as follows:

$$S_{2d} = f * P * R * \{(\bar{S} + A_{id}\alpha_{id} + A_{exp}\alpha_{exp}) + t_{3d}\} \quad (2)$$

where $S_{2d}$ is the 2D coordinate matrix of the 3D face after Weak Perspective Projection, rotation and translation transformation. $f$ is the scaling factor. P is a perspective projection matrix $\begin{pmatrix} 1 & 0 & 0 \\ 0 & 1 & 0 \end{pmatrix}$. R is a rotation matrix constructed according to three rotation angles of *pitch, yaw* and *roll*. $t_{3d}$ is the translation transformation matrix of 3D points. Therefore, for the modeling of a specific face, we only need to solve the 3D parameter $P = [f, pitch, yaw, roll, t_{3d}, \alpha_{id}, \alpha_{exp}]$. Figure.2 shows a 3D face model, which is projected through a weak perspective and aligned with the face model after the original image.

### 3.2. 3D Morphable Model Fitting

The 3DMM fitting process can be expressed as minimizing the error between the 2D coordinates $S_{2d}$ of the 3D point projection and the ground truth $S_{GT}$. Therefore, the parameter $P$ when the error converges to a minimum. The fitting process is expressed as follows:

$$\min_{f, R, t_{3d}, \alpha_{id}, \alpha_{exp}} \|S_{2d} - S_{GT}\| \quad (3)$$

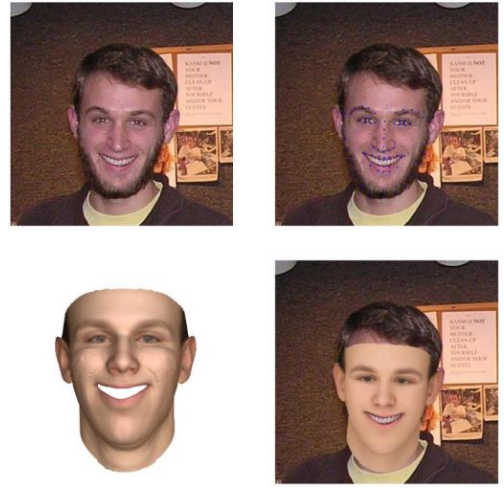

Figure.2. As shown in the top row, we showed the original image and landmark, the second row, from left to right, is the output of the 3D face model and 3D alignment to the 2D image.

Since the face generated by the 3D model has about 50,000 vertices, when iteratively calculating the formula (3), the convergence speed is slow, and the convergence effect is not robust. Therefore, in the fitting process, the landmark(eg, eye center, mouth corner, and nose tip) are used as the ground truth . the error of 3D face feature points of the reconstructed model are minimized. Only calculating landmark makes the model converge faster, while reducing the amount of computation. Therefore, the fitting process of this 3DMM is expressed as follows:

$$\min_{f, R, t_{3d}, \alpha_{id}, \alpha_{exp}} \|S_{2d} - S_{GT}\|_{landmark} \quad (4)$$

Because each part of landmark on 2D image and landmark on 3D model has special semantic information. Therefore the error in the fitting process is different for each landmark. In order to understand these effects, we must analyze the semantic information of each part of the landmark corresponding to 2D-3D. Therefore, we use a weighted landmark 3DMM fitting. Thanks for [4,25,26] research progress in face reconstruction and face alignment. The fitting process of 3DMM can be further expressed as follows:

$$\min_{f, R, t_{3d}, \alpha_{id}, \alpha_{exp}} \|W(S_{2d} - S_{GT})\|_{landmark}^2 \quad (5)$$

where $W$ is the weight matrix of the landmark. To constrain the reconstruction error at each landmark point, we define the weight matrix:

$$W = diag(w_1, w_2, \ldots, w_i, \ldots, w_{n-1}, w_n)$$

$w_i$ represents the error weight of the *i*-th 2D landmark and the corresponding 3D landmark. That is, the larger the error, the larger the weight, and the program preferentially fits the landmark point with a larger weight. This can solve the problem of large local landmark error and enhance the overall reconstruction effect.

### 3.3. Lage Pose Face Reconstruction

The frontal image of the face has more identity information than the profile image, so many existing face recognition methods are based on small and medium poses. Recent research [21] has shown that face recognition using images of frontal faces image is more accurate than profile face image.

In 3D face reconstruction, more focus was on small and medium poses, because some landmark points will become invisible in large poses. This will bring some error to 3D face reconstruction. The error is mainly from the shape information. In order to solve this problem, we use HPEN [21] to normalize the face image in the large poses. HPEN is based on the normalization of the pose and expression of the 3DMM, and Poisson Editting [31] is used to recover the occluded area of the face due to the angle. This method can restore the identity information of the face to a greater extent. We only use the normalization of the pose in this method, and do not normalize the expression. Because 3D face reconstruction requires simultaneous reconstruction of shape and expression information. We define this process as follows:

$$I_f = HPEN(I_o, landmark_{I_o}) \qquad (6)$$

where $I_o$ is the original image, $landmark_{I_o}$ is the landmark corresponding to the original image $I_o$, and $I_f$ is the generated front image. After generating a frontal image, use OpenCV to perform face landmark detection on $I_f$ to generate $landmark_{I_f}$. Then we fit the $I_o$ and $I_f$ separately using equation (5):

$$loss_1, P_1 = Fitting(I_f, landmark_{I_f}) \qquad (7)$$

$$loss_2, P_2 = Fitting(I_o, landmark_{I_o}) \qquad (8)$$

where $loss_1$, $loss_2$ respectively represent the loss value of $I_f$ and $I_o$ after fitting, $P_1, P_2$ respectively represent the 3D parameters that $I_f$ and $I_o$ output after fitting.

Since the front image retains more identity information, and the original image retains more pose and expression information, we weighted sum of the 3D parameters of the two images:

$$P = w_1 * P_1 + w_2 * P_2 \qquad (9)$$

where

$$w_1 = 1 - \{loss_1 / (loss_1 + loss_2)\}$$
$$w_2 = 1 - \{loss_2 / (loss_1 + loss_2)\}$$

where $w_1$, $w_2$ correspond to the weights of the 3D parameters of the front image and the source image, respectively. Since the loss value indicates the degree of precision of the current image during the fitting process. Therefore, the larger the loss, the larger the fitting error; the smaller the loss, the smaller the fitting error, and the more complete the information represented by the 3D parameter. When the loss is smaller, the corresponding weight should be larger. In the large pose, the front image has a smaller fitting error than the source image, and the loss value is relatively small, so our method has better reconstruction performance under large posture.

The 3D parameter $[\alpha_{id}, \alpha_{exp}]$ can be input into the Equation 1 to obtain the 3D face model S. According to Equation 2, the 3D face model S and the transformation parameters $[f, R, t_{3d}]$ are the result of alignment to the face image. See Figure.1 for the framework of our method.

## 4. Experiments and Analysis

In this section, we evaluate the performance of PIFR on three common face reconstruction tasks, face reconstruction in small and medium poses, face reconstruction in large poses, and face reconstruction in extreme poses (±90 yaw angles).

### 4.1. Databases

We evaluate the performance of PIFR on three publicly available face data sets AFW [27], LFPW [28], and AFLW [29]. These three data sets contain small and medium poses, large poses and extreme poses (±90 yaw angles). We divide the dataset AFW and LFPW into three intervals of $[0°, 30°], [30°, 60°], [60°, 90°]$ according to the face absolute yaw angle, and each interval is about 1/3 of the total. The face reconstruction performance evaluation is conducted in small and medium poses AFW and LFPW, and performance evaluation in extreme pose on AFLW.

**AFW.** The AFW dataset is a face image library created using Flickr images, containing 205 images, including 468 marked faces, which contain complex background changes and face pose changes. For each face, there is a rectangular bounding box, 6 landmarks.

**LFPW.** LFPW face database includes two directories, testset and trainset. Testset is a test sample set, including 224 face images, and each is marked with 68 feature points, stored in the .pts file. The trainset directory is a training sample set that includes 811 face images and is labeled with 68 feature points. we selected 900 face images from two sample sets for testing.

**AFLW.** AFLW face database is a large-scale face database including multi-pose and multi-view, and each face is marked with 21 feature points. This database has a very large amount of information, including pictures of various poses, expressions, lighting, and ethnicity. The AFLW face database consists of approximately 250 million hand-labeled face images, of which 59% are women and 41% are men. Most of the image are colored, and only a few are gray image. we only use part of the extreme pose face image of the AFLW database for qualitative analysis.

### 4.2. Experimental Setup

The 3D Morphable Model we use is the BFM model, and the expression model we use the FaceWarehouse [24] model. The 3D model parameter P is a 235-dimensional feature vector, including a 7-dimensional pose parameter $[f, pitch, yaw, roll, t_{3dx}, t_{3dy}, t_{3dz}]$, a 199-dimensional identity parameter $\alpha_{id}$, and a 29-dimensional expression parameter $\alpha_{exp}$.

### 4.3. Evaluation Metric

Given the ground truth 2D landmarks $U_{ij}^g$ and $V_{ij}^r$ estimated landmarks of $N$ testing images. We use Mean Euclidean Metric (MEM) to calculate the error between the estimated and ground truth. In 3D face reconstruction, our goal is to minimize MEM. MEM is defined as follows:

$$\text{MEM} = \sqrt{\frac{1}{N}\sum_{i=1}^{N}\sum_{j}^{K}\left\|U_{ij}^g - V_{ij}^r\right\|^2} \quad (10)$$

where K=68 is the number of landmarks for each face. $U_{ij}^g$ and $V_{ij}^r$ refer to the ground truth and estimated of the $j$th landmark of the $i$th sample.

### 4.4. Comparison Experiments

We present a quantitative comparison of the proposed methods with E-3DMM and FW-3DMM [30] in the datasets AFW and LFPW, and qualitative comparisons on the AFLW dataset. E-3DMM is fitted using Equation 4, and FW-3DMM is fitted using Equation 5, which improves the fitting accuracy of each landmark.

**Comparison on AFW.** In AFW, we estimated the average MEM of 240 samples under the same conditions using E-3DMM, FW-3DMM and PIFR, respectively, with an average of 80 samples per interval. As shown in Tab. 1, Our PIFR shows superior performance compared to the other two methods. The reconstruction accuracy has different degrees of improvement in any poses, especially in the large posture. Fig.3 shows the distribution of MEM in different postures of PIFR, E-3DMM and FW-3DMM. In Figure.3, PIFR is shown to be significantly improved for reconstruction performance in large poses. Fig4 shows the corresponding cumulative errors distribution (CED) curve, which proves that our method has significantly improved the performance of 3D face reconstruction for arbitrary poses and expressions. Fig.8 shows the reconstruction results of our method on AFW

**Comparison on LFPW.** In LFPW, we compute the average MEM of 1190 samples under the same conditions. There are 534 samples in $[0°, 30°]$, 343 samples in $[30°, 60°]$ and 232 samples in $[60°, 90°]$. The results are show in Table.2, the CED curves are plotted in Figure. 5 and different poses of MEM distribution are plotted in Figure.6. The reconstruction results of our method on LFPW are shown in Figure.9.

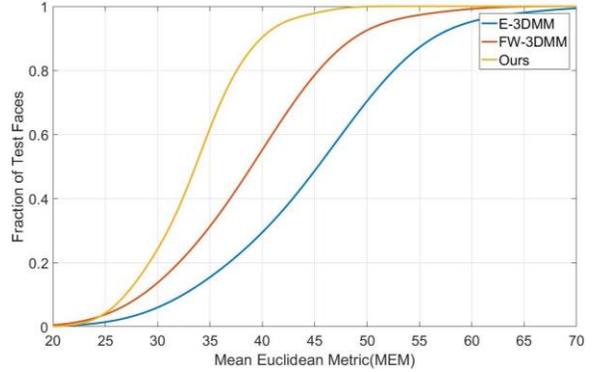

Figure.3. Comparisons of cumulative errors distribution (CED) curves on AFW.

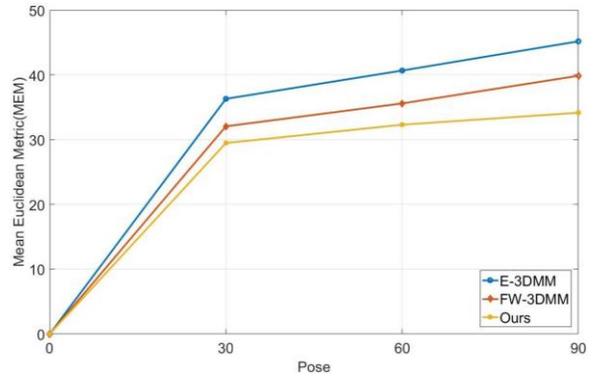

Figure.4. As show the distribution of MEM in different poses of PIFR, E-3DMM and FW-3DMM

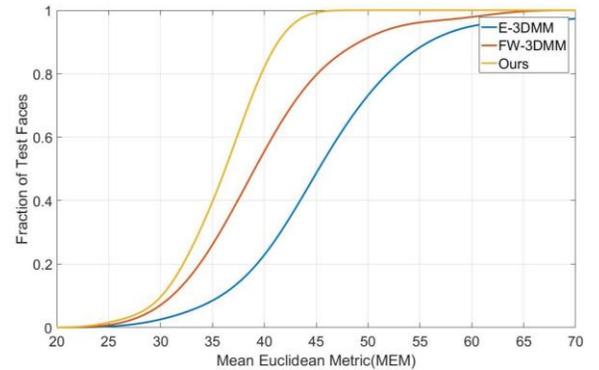

Figure.5. Comparisons of cumulative errors distribution (CED) curves on LFPW

**Comparison on AFLW.** In the AFLW database, we select some face images in extreme poses for qualitative testing. This result is shown in Fig.7. Almost half of the landmarks are invisible due to extreme posture. Therefore, in this case, E-3DMM and FW-3DMM will produce large errors or even failures. Our approach demonstrates superior performance in this situation.

Table.1. The MEM of face reconstruction results on AFW with the first best results highlighted

| Method | AFW(68_pts) | | | | | | | | | | |
|---|---|---|---|---|---|---|---|---|---|---|---|
| | [0,30] | | | [30,60] | | | [60,90] | | | | |
| | Shape | Expression | Sum | Shape | Expression | Sum | Shape | Expression | Sum | Mean | Std |
| E-3DMM | 20.06 | 20.16 | 40.22 | 22.62 | 22.71 | 45.33 | 24.57 | 24.68 | 49.24 | 44.93 | 4.52 |
| FW-3DMM | 17.14 | 17.27 | 36.11 | 19.51 | 19.59 | 39.11 | 21.54 | 21.63 | 43.16 | 39.46 | 3.54 |
| Ours | **15.56** | **16.14** | **31.7** | **16.42** | **16.5** | **32.93** | **17.99** | **18.11** | **36.1** | **33.57** | **2.27** |

Table.2. The MEM of face reconstruction results on LFPW with the first best results highlighted

| Method | LFPW(68_pts) | | | | | | | | | | |
|---|---|---|---|---|---|---|---|---|---|---|---|
| | [0,30] | | | [30,60] | | | [60,90] | | | | |
| | Shape | Expression | Sum | Shape | Expression | Sum | Shape | Expression | Sum | Mean | Std |
| E-3DMM | 18.10 | 18.20 | 36.30 | 20.29 | 20.38 | 40.67 | 22.55 | 22.63 | 45.18 | 40.72 | 4.44 |
| FW-3DMM | 15.98 | 16.08 | 32.06 | 17.75 | 17.85 | 35.59 | 19.88 | 19.98 | 39.86 | 35.84 | 3.91 |
| Ours | **14.69** | **14.79** | **29.48** | **16.09** | **16.22** | **32.31** | **17.02** | **17.15** | **34.16** | **31.98** | **2.36** |

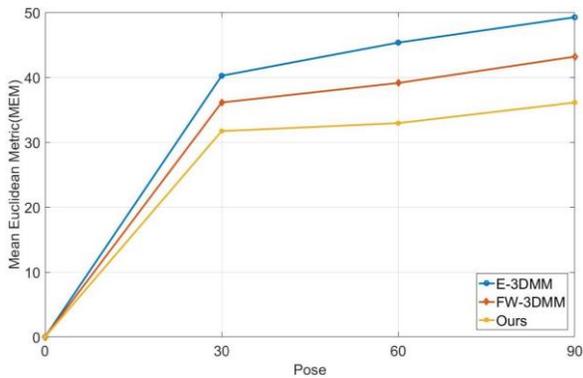

Figure.6. The distribution of MEM in different poses of LFPW

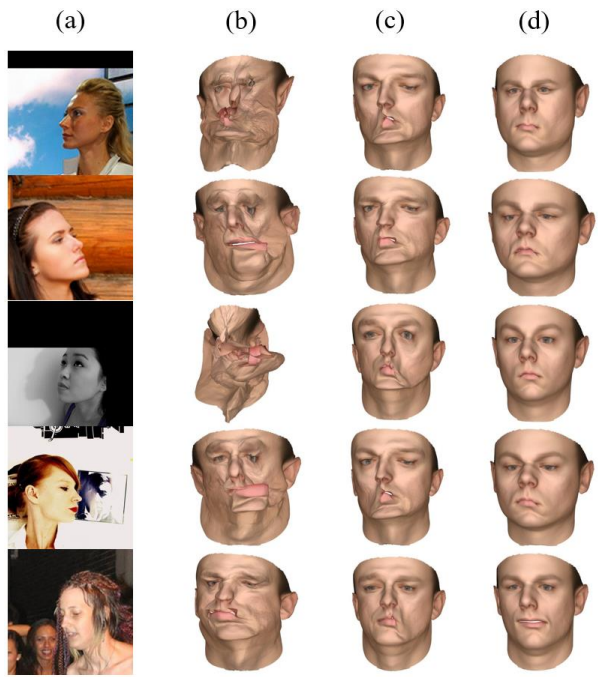

Figure.7. An example of a 3D surface reconstructed on the AFLW dataset by different methods: (a) The input 2D image, (b)FW-3DMM, (c)E-3DMM, (d) Ours

## 5. Conclusion

In this paper, we propose a novel 3D face reconstruction framework PIFR, which solves the problem of 3D face reconstruction under large poses. We have obtained good reconstruction performance even in extreme poses. Our innovative use of positive face images and source images for weighted fusion restores enough face information. Experiments show that our algorithm has excellent reconstruction performance on AFW, LFPW and AFLW. In future, we will further restore more facial identity information in the large poses, and will further improve the accuracy of reconstruction.

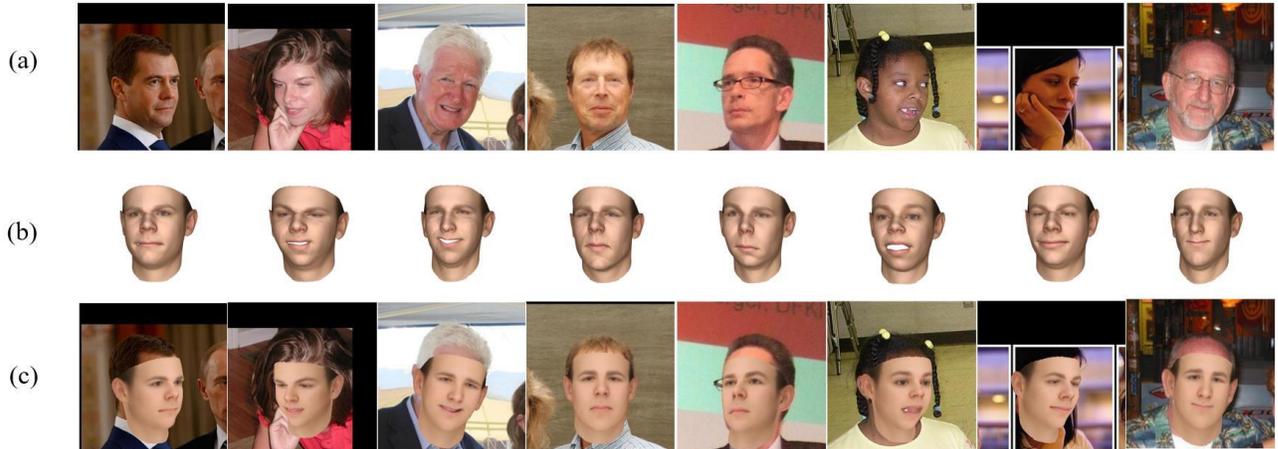

Figure.8. Reconstruction of 3D face for images in AFW by the proposed method: (a)The input 2D image, (b)3D face, (c)Align to 2D image

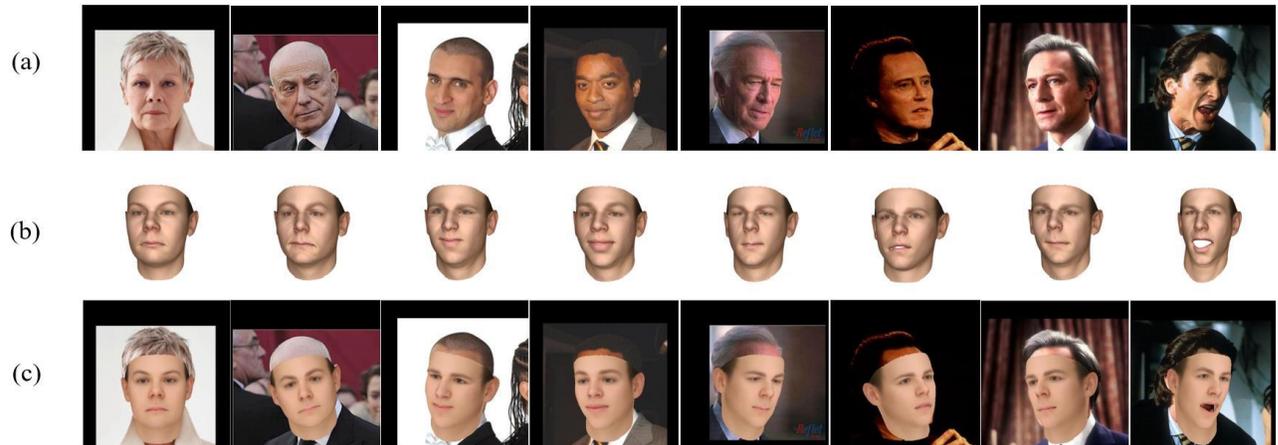

Figure.9. Reconstruction of 3D face for images in LFPW by the proposed method: (a)The input 2D image, (b)3D face, (c)Align to 2D image